\documentclass{article} 
\usepackage{iclr2024_conference,times}

\usepackage{amsmath,amsfonts,bm}

\def\eqref#1{equation~\ref{#1}}

\def\1{\bm{1}}

\DeclareMathAlphabet{\mathsfit}{\encodingdefault}{\sfdefault}{m}{sl}
\SetMathAlphabet{\mathsfit}{bold}{\encodingdefault}{\sfdefault}{bx}{n}

\usepackage{hyperref}
\usepackage{url}
\usepackage{graphicx}
\usepackage{wrapfig}
\usepackage[all]{xy}
\usepackage{tabularx} 

\usepackage[export]{adjustbox}
\usepackage{amsmath,amssymb,amsfonts}
\usepackage{algorithmic}
\usepackage{textcomp}
\usepackage{xcolor}
\usepackage[capitalise,noabbrev]{cleveref} 
\usepackage{csvsimple}
\usepackage{caption}
\usepackage{float}
\usepackage{subcaption}
\usepackage{array}
\usepackage{multirow}

\title{Model Failure or Data Corruption? Exploring Inconsistencies in Building Energy Ratings with Self-Supervised Contrastive Learning}

\author{Qian Xiao, Dan Liu \& Kevin Credit\\
ADAPT Centre\\
O'Reilly Institute, Trinity College 25 Westland Row, Dublin, D02 PN40\\
\texttt{\{qian.xiao, dan.liu, kevin.credit\}@adaptcentre.ie}
}
\iclrfinalcopy 
\begin{document}

\maketitle

\begin{abstract}
Building Energy Rating (BER) stands as a pivotal metric, enabling building owners, policymakers, and urban planners to understand the energy-saving potential through improving building energy efficiency. As such, enhancing buildings' BER levels is expected to directly contribute to the reduction of carbon emissions and promote climate improvement. Nonetheless, the BER assessment process is vulnerable to missing and inaccurate measurements. In this study, we introduce \texttt{CLEAR}, a data-driven approach designed to scrutinize the inconsistencies in BER assessments through self-supervised contrastive learning. We validated the effectiveness of \texttt{CLEAR} using a dataset representing Irish building stocks. Our experiments uncovered evidence of inconsistent BER assessments, highlighting measurement data corruption within this real-world dataset.
\end{abstract}

\section{Introduction}
The building sector is responsible for nearly 40\% of energy-related CO2 emissions \citep{review2022advancing}.
Improving building energy efﬁciency, especially for existing ones, is key to combating climate change for many countries \citep{egan2023building}.  Building energy efficiency assessment plays a pivotal role in guiding decisions for both retrofitting existing buildings and designing new ones \citep{coyne2021mind,energyenergy}. The assessment results stand as cornerstone metrics in shaping various policies for guiding the public to address energy poverty and steer climate action. For example, in Europe, buildings’ energy efficiency is factored into mortgage interests rates for the homeowners \citep{billio2022mortgage}. 
New dwellings in many EU countries are also required to comply with strict energy efficiency regulations \citep{energyenergy}.
Despite its significance, the assessment process of building energy efficiency in which professionals involved is susceptible to missing values and faulty measurements as the result of negligent errors \citep{ucd2020data}. Consequently, such processes are prone to partial data corruption, leading to potential substantial inaccuracies in the assigned energy ratings.

Ensuring the reliability and transparency of the Building Energy Rating (BER) process is crucial for upholding the integrity of energy efficiency assessments. To tackle this challenge, previous research on Ireland's BER data has emphasized the potential benefits of integrating data-driven approaches \citep{ucd2020data}. While such approach is promising, the machine learning models examined in \citep{ucd2020data} exhibited a significant performance drop when predicting fine-grained rating level categories. Table \ref{table:comparison} presents such performance drops observed in both the Random Forest model and the MLP (multi-layer perceptron) deep learning model. Various factors could potentially contribute to these performance drops, including models' limited generalizability, sparsity of data distribution, or poor data quality. Currently, there is still a lack of understanding of the actual reasons behind such observations, which, in turn, hamper the potential of data-driven approaches to facilitate the BER assessment process.

\begin{table}[h!]
\centering
\begin{tabular}{c|c|c|c}
\hline
BER Levels & 
Performance Measure & Random Forest 
& MLP \\ \hline
\multirow{2}{*}{A1, A2, ..., E1, E2, F, G} & Accuracy (\%) & 62.8 & \textbf{69.5} \\  & Macro F1 (\%) & 63.1 & \textbf{63.9} \\ \hline
\multirow{2}{*}{A, B, C, D, EFG}      & Accuracy (\%) & 76.1 & \textbf{88.6} \\ 
                     & Macro F1 (\%) & 75.8 & \textbf{88.9} \\ \hline
\end{tabular}
    \caption{ Performance Comparison of Models for BER Prediction with Different Level Granularity}
\label{table:comparison}
\end{table}
This study aims to investigate such observations using self-supervised deep learning approaches.  
Specifically, we propose \texttt{CLEAR}, a self-supervised learning approach 
that employs Contrastive Learning for Energy Assessment Rating evaluation, aiming to provide explainability and improve the reliability of assessment data. 
Our approach's training process relies entirely on building measurement data, eliminating the need for assessment labels in training. This approach is preferable for our investigation, because the modelling process doesn't require assessment labels that are subjective to individual assessors' judgment. 
Moreover, our approach utilizes the SCARF model \citep{bahri2021scarf} for contrastive learning, providing an advantage in mitigating the potential presence of data noise in modeling. In the modeling process, potential existing data corruption is leveraged to facilitate the randomized feature corruption process. This, in turn, may even enhance the model's generalizability and result in better representations of buildings.

With a Irish building stocks dataset, we discovered evidence of clear inconsistencies in energy ratings. The results demonstrate that buildings with similar feature values were given very different rating levels for their BER assessments in this dataset. 
By revealing these inconsistencies in building energy assessments, we identified significant faulty values in the building features. Consequently, such data corruption may largely limit the potential of data-driven approaches for BER assessment. This explains the observation that most of these approaches failed to achieve satisfactory predictions when the rating was at a fine-grained level.


\section{Approach}


Our approach \texttt{CLEAR} consists of two steps. First, we adopt self-supervised contrastive learning via SCARF modelling \citep{bahri2021scarf} to extract representations for buildings. For contrastive learning, we generate positive pairs by randomly selecting and corrupting a subset of features. The negative pairs are generated by contrasting them with other records in a batch. After training, we extract the latent representation of each building with SCARF's encoder. Second, we explore the model's latent space to examine the rating inconsistency based on latent representations. Specifically, we visualize latent representations via PCA compression in 2D or 3D space. Close proximity of building representations in the latent space indicates similarities in their measured feature values. We can visually identify rating inconsistencies by observing close representations with different BER rating levels in the compressed PCA space. Once the reference buildings are located, we proceed to calculate the nearest neighboring buildings in the derived high-dimensional latent space.



\section{Experiments}
\paragraph{Identifying Rating Inconsistency in the Latent Space.}

\begin{figure*}[t!]
\captionsetup{justification=centering}
        \begin{subfigure}[b]{0.48\textwidth}
         \centering
         \includegraphics[width=\textwidth, 
         height=4cm
         ,trim={0cm 1cm 3cm 3cm},clip
         ]
        {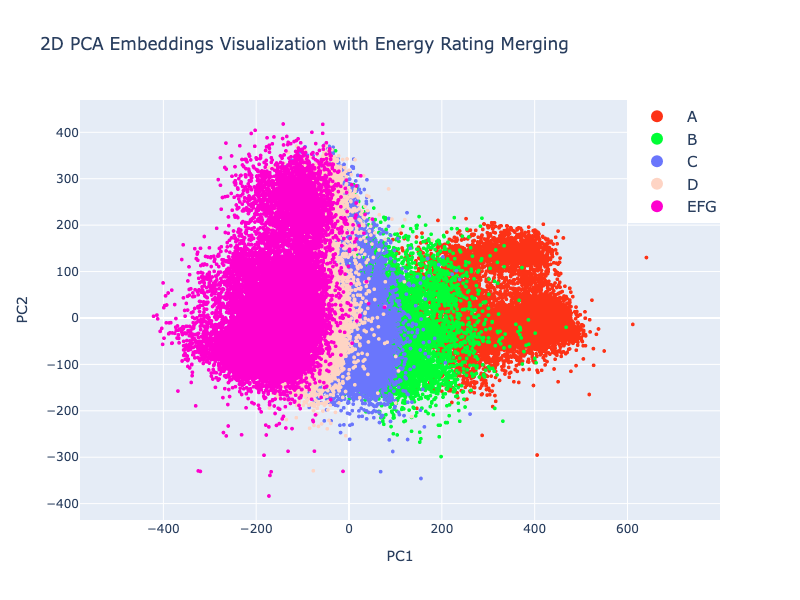}
         \caption{Five BER Levels}
         \label{fig:latent:five}
     \end{subfigure}
     \hfill
               \begin{subfigure}[b]{0.48\textwidth}
         \centering
         \includegraphics[width=\textwidth, 
         height=4cm
         ,trim={0cm 1cm 3cm 3cm},clip
         ]
        {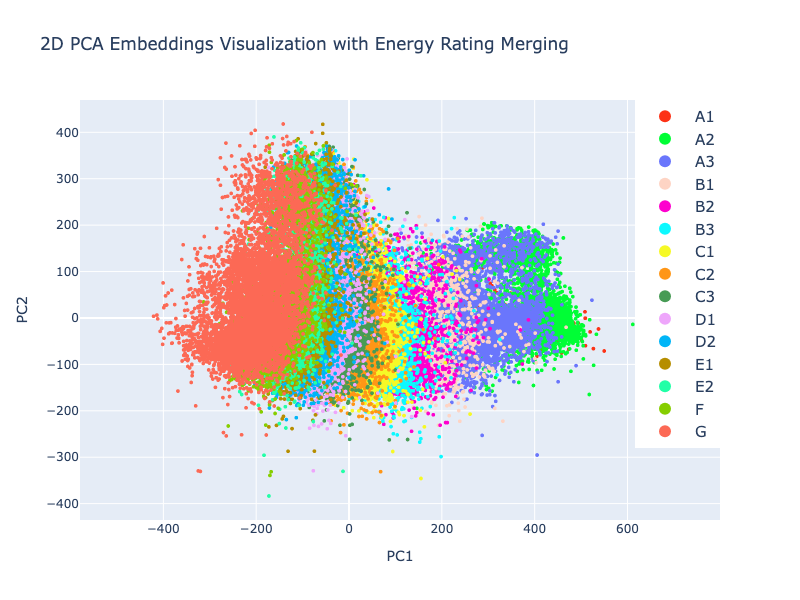}
         \caption{Fifteen BER Levels}
         \label{fig:latent:fifteen}
     \end{subfigure}
         \caption{2D PCA Representations in the Latent Spaces of Scarf Models
        }
        \label{fig:latent}
\end{figure*}

For each building in the dataset\footnote{The description of the dataset we use and our experiments configuration can be found in the appendix.}, we extract the latent representation in a 32-dimensional vector. Subsequently, we derive a 2D visualization by compressing the latent representations further to two dimensions using PCA. \cref{fig:latent} displays the 2D visualization of latent spaces for different BER granularities. As shown in \cref{fig:latent:five}, buildings with five coarse BER levels are spread across the latent space in sequence, aligned with the order of BER levels. Those with the same BER level are predominantly clustered together. However, in \cref{fig:latent:fifteen}, the distribution of latent representations reveals that buildings with adjacent BER levels mix together, and the boundaries of fine-grained rating levels for each coarse level (e.g., `B1', `B2', `B3' for coarse rating `B') are not clear. This indicates inconsistent ratings for buildings with similar feature values, especially evident in neighboring levels such as the group (`B1', `B2', `B3'), the group (`C1', `C2', `C3'), and the group (`D1', `D2'). This observation aligns with the findings in the confusion matrix shown in \cref{fig:confusion} in the appendix for the MLP model.

\begin{figure*}[th!]
\captionsetup{justification=centering}
        \begin{subfigure}[b]{.5\textwidth}
         \centering
         \includegraphics[width=\textwidth, 
         height=5cm
         ,trim={.5cm 2cm 10cm 2cm},clip
         ]
        {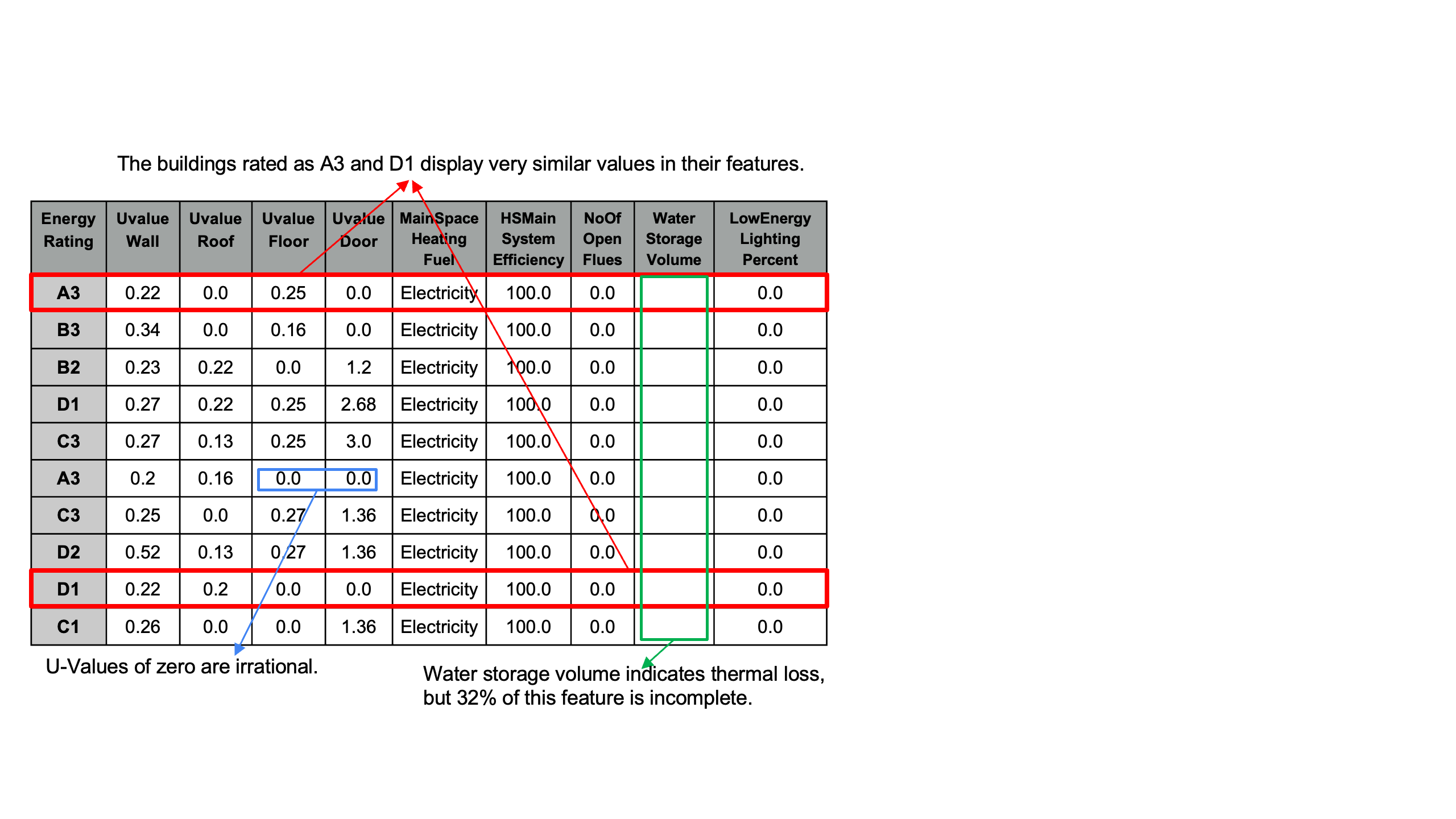}
         \caption{Buildings with BER ranging from A3 to D1}
         \label{fig:a3_to_d1}
     \end{subfigure}
     \hfill
               \begin{subfigure}[b]{.5\textwidth}
         \centering
         \includegraphics[width=\textwidth, 
         height=5cm
         ,trim={1.5cm 3.3cm 9cm 1cm},clip
         ]
        {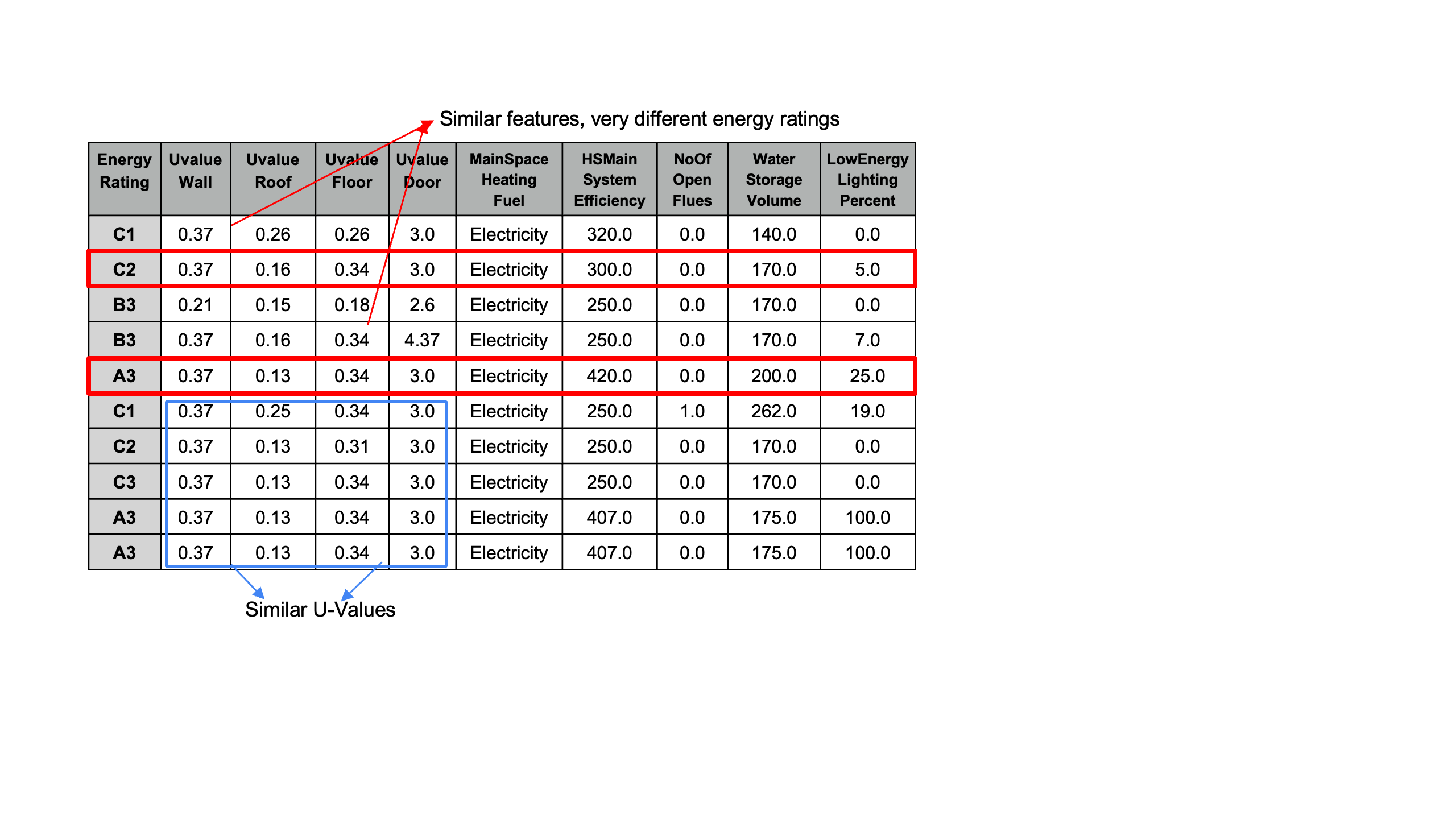}
         \caption{Buildings with BER ranging from A3 to C3}
         \label{fig:a3_to_c3}
     \end{subfigure}
         \caption{Inconsistent Ratings \& Measurement Data Corruption
        }
        \label{fig:inconsistency}
\end{figure*}

\paragraph{Data Corruption in the Inconsistent BER Rating Records.}
The positioning of buildings' representations in the latent space reveals the distribution of buildings with different features. Proximity in the latent space signifies similarity in features between two buildings. Consequently, identifying similar buildings with different BER levels becomes evident in the latent space. For instance, \cref{fig:a3_to_d1} showcases ten buildings resembling the reference building with BER level `A3' (the first row in the table). These buildings are derived by calculating the top ten closest representations in the 32-dimensional latent space. It is notable that these buildings share very similar feature values for U-values of the wall, roof, door, window, and floor—features among those of top importance identified by the decision tree algorithm. Additionally, their heating systems possess identical specifications. However, their ratings range from `A3' to `D1' in the assessment. In addition, we also found out substantial abnormal values in lighting and water storage highlight potential data corruption in their measurements. As another example, \cref{fig:a3_to_d1} presents another group of similar buildings, with a specific focus on two buildings (2nd row and 5th row) that exhibit almost identical feature values across all dimensions but are assessed differently as `C2' and `A3', respectively. The five buildings at the bottom also display similar U-values in all aspects. For reference purpose, details of value ranges for U-values can be presented in the box-plots (see \cref{fig:boxplot} in the appendix).

\section{Conclusion}
In this study, we introduce a data-driven approach based on self-supervised contrastive learning to identify inconsistencies in BER assessments. Using an Irish building stock dataset, our experiments show the effectiveness of the proposed approach in detecting rating inconsistency as well as measurement data corruption in the BER assessment process.

\subsubsection*{Acknowledgments}
This work has received funding from the Science Foundation Ireland's Future Digital Challenge programme for the project entitled ``Exploring realistic pathways to the decarbonization of buildings in the urban context'' under the grant No. 22/NCF/FD/10985.
\bibliography{reference}
\bibliographystyle{iclr2024_conference}

\appendix
\section{Appendix}
\subsection{Data Preprocessing \& Feature Extraction}
The data preprocessing consists of three steps. The first step is data cleaning. In this step, we use the interquartile range (IQR) technique to identify and address outliers. 
We then group the data by building type. For numerical features, missing values are imputed with the mean value specific to each building type. For categorical features, we utilized the most frequent value in each category for fill-ins. In the second step, we selected 40
features from the total 211 features in the original dataset. 
The selection is based on the feature importance analysis with decision tree algorithm, which identified the features with the most significant impact on energy ratings. During this process, we excluded compound features that can be 
derived by the actual energy ratings, such as total primary energy use for the dwelling 
and the dwelling's carbon dioxide emissions (known as CO2 emissions). These 40 features are categorised into distinct groups, each describing
different aspects of a building. These categories include building envelope features (e.g., the area of the wall, roof, and door), building fabric features (e.g., the U-values for the wall, roof, and door), characteristics of the heating system (e.g., the main heating system efficiency), hot water-related attributes (e.g., water storage volume), and spatial features (e.g., the county code). Lastly, in the third step, we applied standard scalers to standardize the numerical features and one-hot encoders to transform categorical features. 

\subsection{Dataset}
We use the Energy
Performance Certificates (EPC) dataset \footnote{https://ndber.seai.ie/BERResearchTool/ber/search.aspx} in this study, collected by the
Sustainable Energy Authority of Ireland. This dataset contains 112,528 building assessment records. There are 15 different BER levels in total, 
ranging from A1 as the highest level to G as the lowest level.

\subsection{Experiments Configuration}

\begin{figure}[th!]
\begin{minipage}[b]{0.5\textwidth}
    \captionsetup{justification=centering}
         \centering
         \includegraphics[width=\textwidth, 
         height=4.5cm
         ,trim={1cm 1cm 1cm 3cm},clip
         ]
        {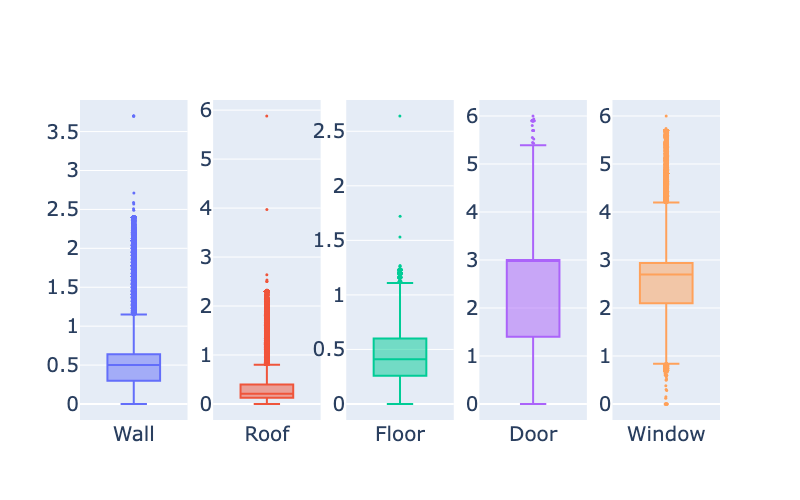}
         \captionof{figure}{Box-plots of U-Values}
         \label{fig:boxplot}
         \end{minipage}
         \hfill
         \begin{minipage}[b]{0.5\textwidth}
             \captionsetup{justification=centering}
         \centering
         \includegraphics[width=.8\textwidth, 
         height=4.5cm
         ,trim={0cm 0cm 0cm 0cm},clip
         ]
        {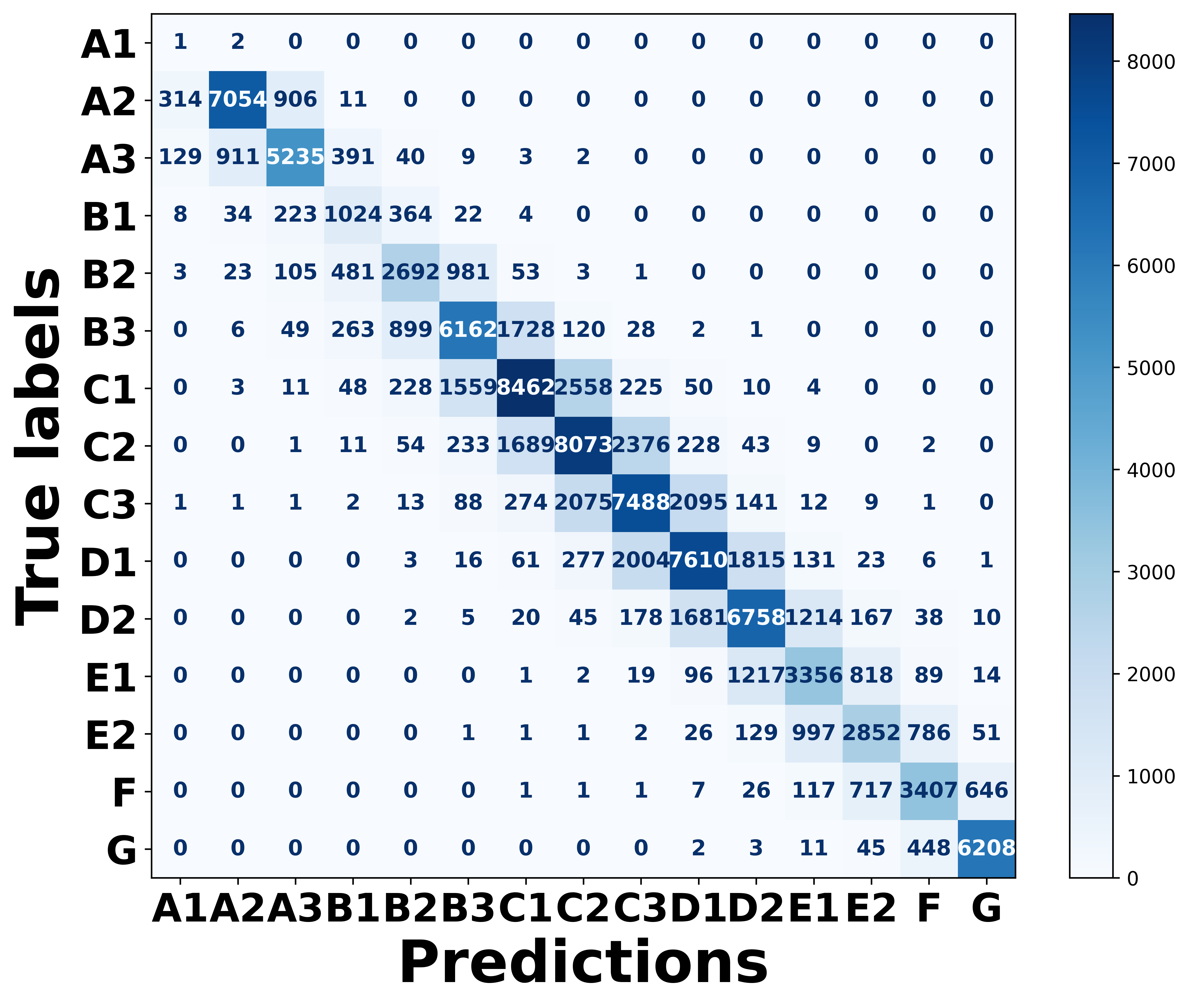}
         \captionof{figure}{Confusion Matrix for MLP Predictions
        }
        \label{fig:confusion}
         \end{minipage}
\end{figure}

In our experiments, the SCARF model was trained for 15 epochs, with a batch size of 16 and a learning rate of 0.001. 
The model used in the experiments consists of three MLP blocks, each with a linear layer followed by ReLU activation. We set the dimension of the model's encoder to be 32. The proportion of randomly selected feature corruption is set to be 30\%. We split the dataset into train, validation, and test sets by the ratio 80\%, 10\%, and 10\%.

\end{document}